\pdfoutput=1
\documentclass[10pt,twocolumn,letterpaper]{article}

\usepackage{amsmath}
\usepackage{amssymb}
\usepackage{cvpr}
\usepackage{epsfig}
\usepackage{graphicx}
\usepackage[breaklinks=true,bookmarks=false]{hyperref}
\usepackage{times}
\usepackage[dvipsnames]{xcolor}

\def\H{1}
\newcommand{\mm}[1]{\if\H1 {\color{Green} [#1]} \fi}
\newcommand{\pp}[1]{\if\H1 {\color{Blue} [#1]} \fi}
\newcommand{\bb}[1]{\if\H1 {\color{Red} [#1]} \fi}

\cvprfinalcopy
\def\cvprPaperID{****}

\begin{document}

\title{HMM-guided frame querying for bandwidth-constrained video search}

\author{
    Bhairav Chidambaram\textsuperscript{1} \\
    {\tt\small bchidamb@caltech.edu}
    \and
    Mason McGill\textsuperscript{1} \\
    {\tt\small mmcgill@caltech.edu}
    \and
    Pietro Perona\textsuperscript{1} \\
    {\tt\small perona@caltech.edu}
}

\maketitle

\begin{abstract}
    We design an agent to search for frames of interest in video stored on a remote server, under bandwidth constraints. Using a convolutional neural network to score individual frames and a hidden Markov model to propogate predictions across frames, our agent accurately identifies temporal regions of interest based on sparse, strategically sampled frames. On a subset of the ImageNet-VID dataset, we demonstrate that using a hidden Markov model to interpolate between frame scores allows requests of 98\% of frames to be omitted, without compromising frame-of-interest classification accuracy.
\end{abstract}

\footnotetext[1]{California Institute of Technology}

\section{Introduction}

We consider the problem of detecting frames of interest in videos stored on a remote server, from which we can request individual frames, at a fixed transmission cost per frame. We define a ``frame of interest'' as one containing at least one object belonging to any member of a set of task-specific object classes.

Our proposed solution is based on decomposing this problem into three constituent subproblems: (1) assigning a score to each observed frame corresponding to the probability, based on observing it in isolation, that it is ``of interest'', (2) integrating this sequence of sparsely observed ``interestingness scores'' into a probability distribution over the full label sequence, and (3) deciding which unobserved frame, if any, to request next.

We solve (1) by learning a convolutional frame-scoring network from ground-truth (frame, label) pairs. We solve (2) using a hidden Markov model (HMM) derived from the transition and co-ocurrence statistics of ground-truth frame labels and regressed frame scores. Our solution to (3) is the following greedy policy: if the application context provides enough time and bandwidth to request another frame, make the frame request that would most lower the mean (across frames) expected cross entropy \cite{murphy2012probabilistic} between frame labels and marginal distributions over them. We evaluate this approach on a subset of the ImageNet-VID dataset \cite{russakovsky2015imagenet}

\begin{figure}[t]
\begin{center}
    \includegraphics[width=0.8\linewidth]{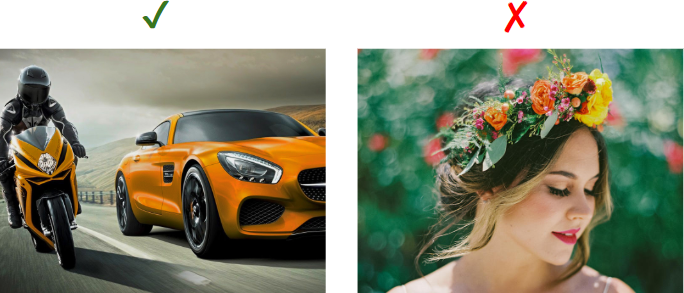}
\end{center}
   \caption{An illustration of frame-of-interest classification. Each frame is assigned a label $\in\{0,1\}$ indicating whether any of the target classes $\mathcal{C}$ are present. For $\mathcal{C} = \lbrace car,\ truck,\ bus \rbrace$, the left frame is labeled $1$ and the right frame is labeled $0$.}
\label{fig:long}
\label{fig:onecol}
\end{figure}

\section{Related work}

The collection of large-scale labeled video datasets like ImageNet-VID \cite{russakovsky2015imagenet} and YouTube-BB \cite{real2017youtubeboundingboxes} has allowed systems for detecting objects in images to be adapted to work with videos (\cite{bertasius2018object}, \cite{Wang2018FullyMN}, \cite{Zhu_2018}). And due to the high computational cost of processing videos, several approaches to reducing system resource demands have been proposed.

Zhu et al. \cite{zhu2018highmob} train separate lightweight networks for (1) detecting objects in sparsely sampled keyframes and (2) estimating motion fields between keyframes, which they use to interpolate between object states inferred from those keyframes. Wang et al. \cite{wang2018fast} exploit motion vectors computed during H.264 compression to efficiently propagate features across frames. In the AdaScale system proposed by Chin et al. \cite{chin2019adascale}, an adaptive agent is trained to select the appropriate scale for the input image, optimizing for speed and accuracy.

Canel et al. \cite{canel2019scaling} propose an on-sensing-platform filtering system to reduce bandwidth consumption in remote video camera deployments. Their micro-classifier predicts the relevance of each frame, and only transmit frames with relevance scores exceeding a threshold. From this work, we borrow the task of frame-of-interest classification. We also take inspiration from their architecture for full-frame object detection, which applies a $1 \times 1$ convolution to the feature map generated by a convolutional feature-extractor and computes the $\max$ over spatial locations.

\section{Approach}

The inference agent we propose maintains a probability distribution over the full label sequence of the video as well as a recommendation for which frame to request next, and updates both as new frames are received. The agent has three components: a convolutional frame-scoring network, a hidden Markov model for extrapolating inference across frames, and a request-recommendation rule (greedy expected-cross-entropy minimization).

\subsection{Scoring retrieved frames}

We use a convolutional frame-scoring network to regress the probability that a given retrieved frame is a frame of interest. We refer to this probability estimate as a ``score'' both because it is uncalibrated \cite{guo2017calibration} and to disambiguate it from probabilities computed using the hidden Markov model.

The frame-scoring network, based on the network described in \cite{canel2019scaling}, is the composition of an ImageNet-pretrained ResNet-18 ``backbone'', with global average pooling and fully connected layers removed \cite{russakovsky2015imagenet,he2016resnet}, a learned $1 \times 1$ convolution with 1 output channel, and a $\mathrm{softmax}$ operation.

\subsection{Inferring dense labels from sparse scores}

\begin{figure}[t]
\begin{center}
    \vspace{0.5cm}    
    \includegraphics[width=0.8\linewidth]{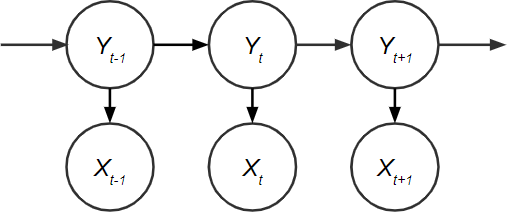}
\end{center}
   \caption{Our hidden Markov model. $Y_t$ corresponds to the ground-truth label of frame $t$ and $X_t$ corresponds to its score.}
\label{hmm}
\end{figure}

\begin{figure}[t]
\begin{center}
    \includegraphics[width=\linewidth]{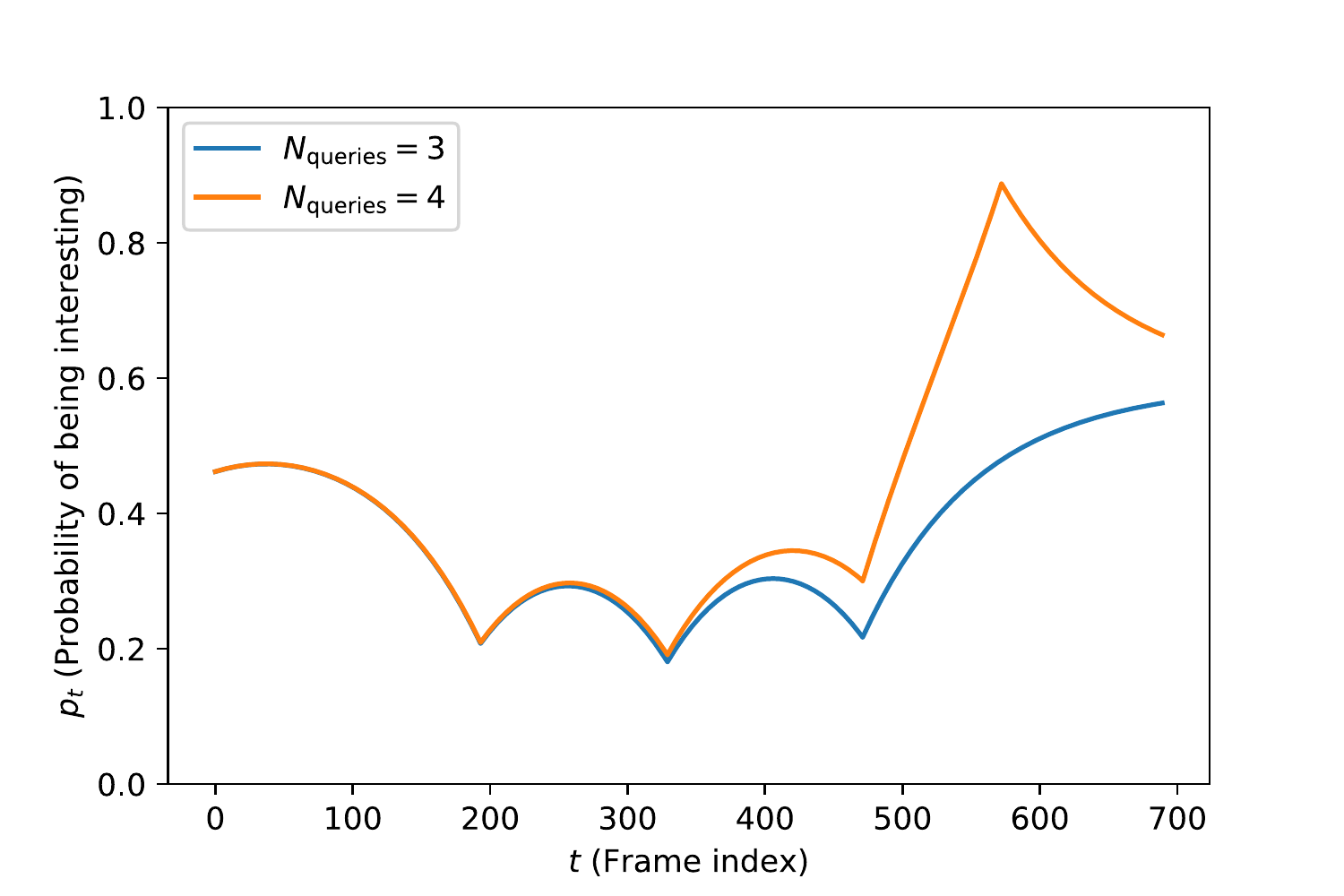}
\end{center}
    \caption{An example marginal frame-label probability vector (in blue), computed based on the results of three queries, and an updated probability vector (in orange) adjusted based on the result of a fourth query.}
\label{ex_belief}
\end{figure}

We construct a hidden Markov model (Fig. \ref{hmm}) to form beliefs over a video's full sequence of frame labels from sparsely observed frame scores. The hidden states take one of $N_Y=2$ values: $1$ if the frame is of interest, and $0$ otherwise. The observed states take one of $N_X = 3$ values, each indicating that the corresponding frame score falls into one of three equiprobable quantiles. We use the forward-backward algorithm \cite{rabiner1986introduction} to compute marginal label probabilities. In Fig. \ref{ex_belief}, we show how $p$ updates according to this model after a new observation.

\subsection{Deciding which frames to request}

While refining our beliefs about a $T$-frame video, we maintain a marginal label probability vector $p \in [0,1]^T$. $p_t = 1$ indicates complete certainty that frame $t$ is of interest, and $p_t = 0$ indicates complete certainty that it is not.

Our agent's goal is to produce a sequence of queries $Q = (Q_1..Q_k)$ such that after observing frames $Q_1, Q_2, ..., Q_k$ and updating $p$ accordingly, it minimizes the mean frame-wise cross entropy $H$ between the distributions specified by $p$ and the ground truth label vector $Y$:
$$
    \frac{1}{T} \sum_{t=1}^T H(Y_t, p_t) =
        - \frac{1}{T} \sum_{t=1}^T {
            Y_t \log p_t
            + (1 - Y_t) \log (1 - p_t)
        }.
$$

Since $Y$ is unknown during inference, we take the expectation over possible outcomes according to $p$. We define the frame-wise expected cross entropy loss $\hat{H}$ as
\begin{align*}
    \hat{H}(p) &= {
        \frac{1}{T} \sum_{t=1}^T {
            \mathbb{E}[H(Y_t, p_t)]
            \text{ with }
            Y_t \sim \operatorname{Bernoulli}(p_t)
        }
    } \\
    &= {
        -\frac{1}{T} \sum_{t=1}^T {
            p_t \log p_t + (1 - p_t) \log (1 - p_t)
        }
    }.
\end{align*}

Using this formula, we compute the expected loss $\hat\ell_q$ of the updated marginal probability vector $p'$ for every possible next query $q \in \{1..T\}$:
\begin{align*}
    \hat\ell_q =\ &\mathbb{E}[\hat{H}(p'|Q_\text{next}=q,X_q=x_q)] \\
           &\text{ with } x_q \sim \operatorname{Emission}(Y_q) \\
           &\text{ and } Y_q \sim \operatorname{Bernoulli}(p_q),
\end{align*}
where $\operatorname{Emission}(y)$ is the hidden Markov model's observation emission distribution for hidden state $y$. We select the query that minimizes this expected loss:
$$
    Q_\text{next} = \operatorname{argmin}_q \hat\ell_q.
$$

\begin{figure*}
\begin{center}
    \vspace{0.5cm}
    \includegraphics[width=\linewidth]{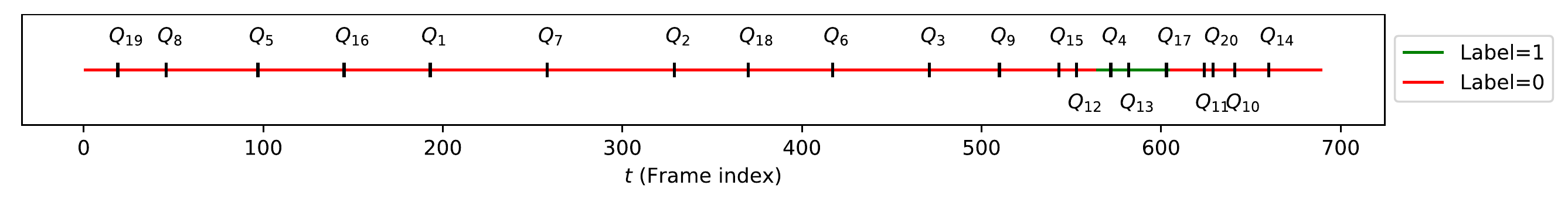}
\end{center}
    \caption{The first few query locations relative to ground truth frame-of-interest labels for an example video. Our agent queries more densely around periods where the label shifts between $0$ and $1$.}
\label{ex_queries}
\end{figure*}

\section{Experiment details}

We test our approach on the task of identifying frames containing road vehicles in the ImageNet-VID dataset, with the target class set $\mathcal{C} = \lbrace bicycle,\ motorcycle,\ car,\ bus \rbrace$. We assign labels to every frame in ImageNet-VID using the provided annotations. If a frame contains an instance of a class in $\mathcal{C}$, it is labeled $1$. Otherwise, it is labeled $0$.

\subsection{Training the frame-scoring network}

The only parameters of the frame-scoring network that we learn are those of the final $1\times 1$ convolution operation. Our training set consists of 10,000 images sampled as follows. We divide all frames in ImageNet-VID into two sets according to their label ($0$ or $1$). From each set, we select 5000 uniformly at random without replacement. This sampling procedure ensures training set diversity. We train the network to predict frame labels using a binary cross-entropy loss. We train for 100 epochs, using the Adam optimizer \cite{kingma2014adam} with a learning rate of $10^{-4}$ and mini-batch size of 8.

\subsection{Computing transition matrices}

The hidden Markov model is parameterized by two matrices: the $2 \times 2$ hidden state transition matrix and the $2 \times 3$ observation emission matrix. We compute these matrices based on the videos in the ImageNet-VID training set that contain at least one frame of interest. The transition matrix is composed of the occurence rates of adjacent label-label pairs. To construct the emission matrix, we first score every frame with the frame-scoring network. We then convert the scores to discrete observations by binning them into three equiprobable quantiles. Finally, we assemble the emission matrix from label-observation co-occurrence rates.

\subsection{Evaluation}

The evaluation dataset consists of the videos in the ImageNet-VID validation set containing at least one frame of interest. To reduce the evaluation run time, videos in this set longer than 300 frames are split into clips of 300 frames or fewer. We define the bandwidth ratio $B$ as the fraction of frames in the video our agent is allowed to observe. \textit{i.e.} our agent can observe $\lfloor BT \rfloor$ frames of a $T$-frame video. After our agent has made this number of observations, we measure frame-of-interest classification accuracy. We measure this accuracy at each bandwidth ratio $B \in \lbrace 0.0, 0.005, ..., 0.1 \rbrace$ and average across videos in our evaluation dataset (Fig. \ref{tradeoff}).

\section{Results}

\begin{figure}[t]
\begin{center}
    \includegraphics[width=\linewidth]{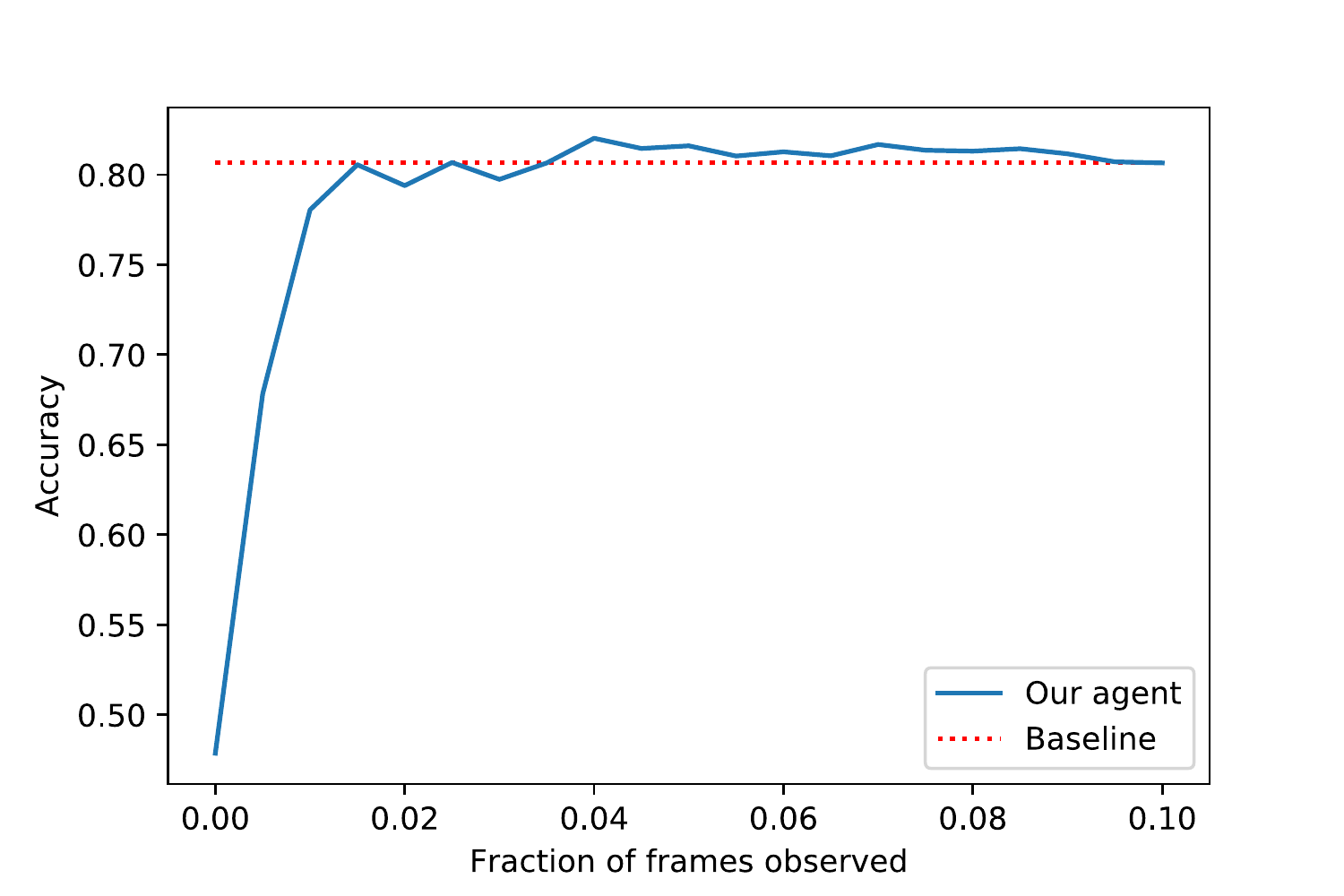}
\end{center}
    \caption{The accuracy-versus-bandwidth curve obtained by evaluating on our validation subset of ImageNet-VID. The dashed red line indicates the accuracy of a full-bandwidth detector, for comparison. Observing as little as two out of every hundred frames, the querying agent is able to make-frame level predictions with accuracy comparable to that of a full-bandwidth detector.}
\label{tradeoff}
\end{figure}

Fig. \ref{tradeoff} displays the accuracy-versus-bandwidth curve measured in our experiment. After observing only 2\% of the frames in a video, our agent is able to make frame-of-interest predictions with the same accuracy as an unconstrained detector. Qualitatively, the agent seems to query densely around ambiguous sections of the video while ignoring others (Fig. \ref{ex_queries}).

\section{Conclusion}

Using a convolutional network to score sparsely observed frames and a hidden Markov model to make dense predictions and generate queries, we can accurately determine which frames in a video contain objects of interest. On a subset of the ImageNet-VID dataset, our method achieves an observation reduction of 98\%.

To improve upon our approach, it may be valuable to further investigate the temporal statistics of our task. Road vehicles are fast-moving objects that only stay in-frame for short periods of time. The statistics of other tasks, like animal- or person-detection, where objects of interest enter and exit less frequently, may be different. The effectiveness of our approach may vary significantly with the statistics of the training and evaluation video sets, and more work is needed to fully characterize this variation.

Furthermore, since we designed our frame-scoring network with simplicity in mind, our system may be improved by substituting in a more sophisticated network. One possibility is to use an architecture incorporating object localization rather than direct, frame-level classification. The downside, however, would be a longer run time and consequently a lower maximum query rate.

\bibliographystyle{ieee_fullname.bst}
{\small \bibliography{egbib.bib}}

\end{document}